%% file: main-arxiv.tex
\documentclass{article}

\usepackage{docmute}

\input{settings}

\title{{\InstantFT}: An FPGA-Based Runtime Subsecond Fine-tuning of CNN Models}

\renewcommand{\shorttitle}{{\InstantFT}: An FPGA-Based Runtime Subsecond Fine-tuning of CNN Models}

\hypersetup{
  pdftitle = {{\InstantFT}: An FPGA-Based Runtime Subsecond Fine-tuning of CNN Models},
  pdfsubject = {cs.LG},
  pdfauthor = {Keisuke Sugiura, Hiroki Matsutani},
  pdfkeywords = {On-device Learning, On-device Fine-tuning, TinyML, Edge Machine Learning, Low-Rank Adaptation, LoRA, FPGA, Hardware Accelerator}
}

\author{%
  \href{https://orcid.org/0000-0001-8534-2381}%
  {\includegraphics[scale=0.06]{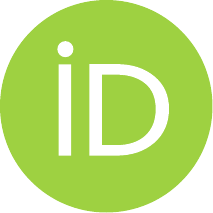}%
  \hspace{1mm}Keisuke Sugiura}\\
  University of Tsukuba\\
  1-1-1 Tenn\^{o}dai, Tsukuba, Ibaraki, Japan\\
  \texttt{sugiura@lila.cs.tsukuba.ac.jp}\\
  \And
  \href{https://orcid.org/0000-0001-9578-3842}%
  {\includegraphics[scale=0.06]{orcid.pdf}%
  \hspace{1mm}Hiroki Matsutani}\\
  Keio University\\
  3-14-1 Hiyoshi, Kohoku-ku, Yokohama, Japan\\
  \texttt{matutani@arc.ics.keio.ac.jp}
}

\begin{document}

\maketitle

\begin{abstract}
Training deep neural networks (DNNs) requires significantly more computation and memory than inference, making runtime adaptation of DNNs challenging on resource-limited IoT platforms.
We propose {\InstantFT}, an FPGA-based method for ultra-fast CNN fine-tuning on IoT devices, by optimizing the forward and backward computations in parameter-efficient fine-tuning (PEFT).
Experiments on datasets with concept drift demonstrate that {\InstantFT} fine-tunes a pre-trained CNN 17.4x faster than existing Low-Rank Adaptation (LoRA)-based approaches, while achieving comparable accuracy.
Our FPGA-based {\InstantFT} reduces the fine-tuning time to just 0.36s and improves energy-efficiency by 16.3x, enabling on-the-fly adaptation of CNNs to non-stationary data distributions.
\end{abstract}

\maketitle

\section{Introduction} \label{sec:intro}
\begin{figure}
  \centering
  \includegraphics[keepaspectratio, width=0.35\linewidth]{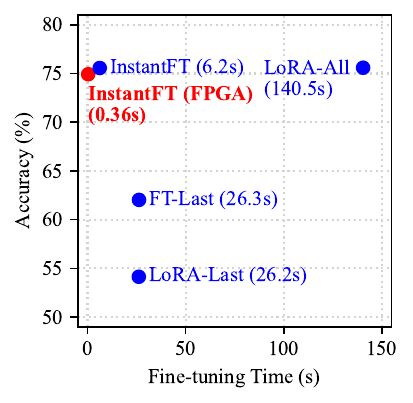}
  \caption{Fine-tuning time vs. accuracy of {\InstantFT} and baselines (Rotated Fashion-MNIST, 75deg).}
  \label{fig:time-vs-accuracy}
\end{figure}

On-device learning of DNNs is an effective solution to bridge the gap between training data and deployed environment~\cite{HanCai20}.
Since full retraining requires prohibitive costs, lightweight PEFT methods that only update a small set of parameters have been explored for resource-limited edge devices~\cite{KazukiSunaga23,HirokiMatsutani25}.
LoRA (Low-Rank Adaptation)~\cite{EdwardJHu22,TimDettmers23} introduces trainable low-rank matrices while keeping the original pre-trained network fixed, allowing efficient adaptation to new tasks with minimal overhead.
Instead of backpropagation and gradient descent, other approaches, e.g., Extreme Learning Machines (ELMs)~\cite{NanyingLiang06} and zeroth-order optimization (ZOO)~\cite{SijiaLiu20}, have been applied to on-device learning.
However, ELM is only applicable to tiny-scale models consisting of one hidden layer~\cite{MinetoTsukada20}, while ZOO suffers from a slow convergence and long training time~\cite{YequanZhao24}.
This work presents {\InstantFT}, an ultra-fast fine-tuning approach based on LoRA.
As shown in Fig. \ref{fig:time-vs-accuracy}, {\InstantFT} runs significantly faster than the baseline LoRA approaches, without compromising accuracy and robustness against data distribution shifts.

\section{Baselines} \label{sec:baselines}
\begin{figure*}
  \centering
  \includegraphics[keepaspectratio, width=\linewidth]{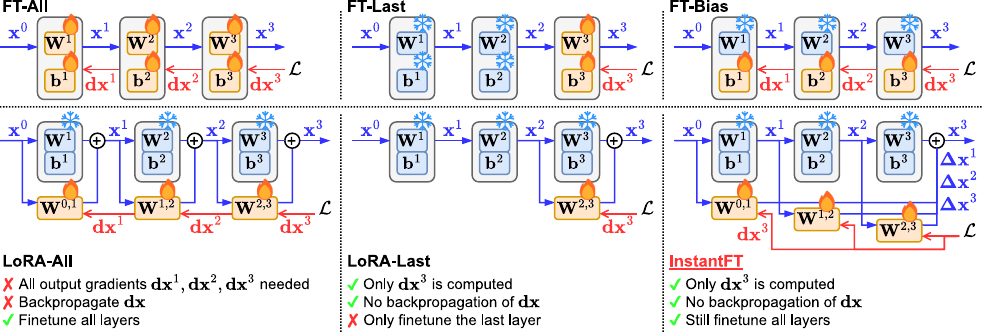}
  \caption{Baseline fine-tuning methods and {\InstantFT}.}
  \label{fig:finetune-approaches}
\end{figure*}

This section describes the baseline fine-tuning strategies.
Table \ref{tbl:comparison} presents a quantitative comparison between our {\InstantFT} and baselines when applied to a LeNet-5-like model (Fig. \ref{fig:cnn-model}) for two image classification datasets.
We consider a simple $L$-layer network composed of fully-connected (FC) and convolution (Conv) layers, each having a weight and bias $\left\{ \mathbf{W}^i, \vb{b}^i \right\}_{i = 1, \ldots, L}$.
We denote the forward activation of the $i$-th layer by $\vb{x}^i$.
The FC layer computes $\vb{x}^i = \mathbf{W}^i \vb{x}^{i - 1} + \vb{b}^i$, where $\vb{x}^{i - 1}, \vb{x}^i, \vb{b}^i \in \mathbb{R}^d, \mathbf{W}^i \in \mathbb{R}^{d \times d}$.
The simplest approach is to fine-tune the entire network (\textbf{FT-All}) or only biases $\left\{ \vb{b}^1, \ldots, \vb{b}^L \right\}$ (\textbf{FT-Bias}~\cite{HaoyuRen21}).
As illustrated in Fig. \ref{fig:finetune-approaches} (top right), another possible approach is to fine-tune parameters of the last layer $(\mathbf{W}^L, \vb{b}^L)$ while keeping the rest frozen (\textbf{FT-Last}).
While FT-Last requires significantly (854--1133x) less computational cost for backpropagation than FT-All, it only updates 1.4\% of the parameters and has a limited fine-tuning capability (Fig. \ref{fig:time-vs-accuracy}).
FT-Bias updates even fewer parameters than FT-Last, but it only reduces the backward FLOPs by 2.4--3.1x compared to FT-All, as it still requires gradients of forward activations $\vb{dx}$.

\begin{table*}[ht]
  \centering
  \caption{Comparison between {\InstantFT} and baselines (LeNet-5)}
  \label{tbl:comparison}
  \setlength{\tabcolsep}{3pt}
  \begin{tabular}{l|rrrr|rrrr} \hline
    & \multicolumn{4}{c|}{MNIST} & \multicolumn{4}{c}{SVHN} \\
    Method & \makecell{Trainable \\ params} &
      \makecell{FLOPs \\ (forward)} & \makecell{FLOPs \\ (backward)} &
      \makecell{Memory \\ usage (KB)} &
      \makecell{Trainable \\ params} &
      \makecell{FLOPs \\ (forward)} & \makecell{FLOPs \\ (backward)} &
      \makecell{Memory \\ usage (KB)} \\ \hline
    FT-All & 61706 & 0.851M & 1.444M & 567.8KB &
      62006 & 1.321M & 1.914M & 579.4KB \\
    FT-Last & 850 & 0.851M & 0.002M & 285.8KB &
      850 & 1.321M & 0.002M & 296.1KB \\
    FT-Bias & 236 & 0.851M & 0.611M & 322.0KB &
      236 & 1.321M & 0.611M & 332.3KB \\
    LoRA-All & 36328 & 0.923M & 0.743M & 611.8KB &
      45480 & 1.412M & 0.762M & 695.4KB \\
    LoRA-Last & 376 & 0.852M & 0.001M & 285.4KB &
      376 & 1.322M & 0.001M & 295.8KB \\
    \textbf{\InstantFT} & 10456 & 0.872M & 0.036M & 366.2KB &
      19608 & 1.360M & 0.054M & 449.8KB \\ \hline
  \end{tabular}
\end{table*}

As in Fig. \ref{fig:finetune-approaches} (bottom), we consider two LoRA-based approaches: inserting LoRA adapters into all layers (\textbf{LoRA-All}) or applying LoRA to only the last layer (\textbf{LoRA-Last}).
For the $i$-th layer, LoRA introduces a new matrix $\mathbf{W}^{i - 1, i} \in \mathbb{R}^{d \times d}$.
It is further decomposed into two trainable matrices $\mathbf{A} \in \mathbb{R}^{r \times d}, \mathbf{B} \in \mathbb{R}^{d \times r}$ of rank $r \ll d$, which are initialized with random Gaussian values and zeros.
The modified layer now produces:
\begin{align*}
  \vb{x}^i &= \left( \mathbf{W}^i + \mathbf{W}^{i - 1, i} \right) \vb{x}^{i - 1} + \vb{b}^i
    = \left( \mathbf{W}^i + \mathbf{B}^{i - 1, i} \mathbf{A}^{i - 1, i} \right) \vb{x}^{i - 1} + \vb{b}^i.
\end{align*}
The trainable parameters in LoRA-All and LoRA-Last are denoted as $\left\{ \mathbf{A}^{i - 1, i}, \mathbf{B}^{i - 1, i} \right\}_{i = 1, \ldots, L}$ and $\left\{ \mathbf{A}^{L - 1, L}, \mathbf{B}^{L - 1, L} \right\}$, respectively.
LoRA-All performs backpropagation across the entire network, and therefore needs to store additional LoRA parameters, all forward activations $\vb{x}^1, \ldots, \vb{x}^L$, and their gradients $\vb{dx}^1, \ldots, \vb{dx}^L$, resulting in 7.7--20.0\% higher memory consumption than FT-All.
Similar to FT-Bias, LoRA-All still retains 39.8--51.5\% of the backward FLOPs of FT-All due to the computation of activation gradients.
In contrast, LoRA-Last updates only the last layer using the activation $\vb{x}^L$, resulting in significantly reduced FLOPs and memory costs for backpropagation, at the cost of lower fine-tuning capability (Fig. \ref{fig:time-vs-accuracy}).

\begin{figure*}
  \centering
  \includegraphics[keepaspectratio, width=0.9\linewidth]{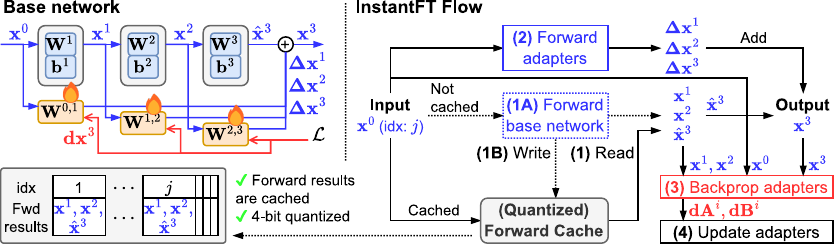}
  \caption{Flow of fine-tuning with {\InstantFT}.}
  \label{fig:instantft-flow}
\end{figure*}

\section{{\InstantFT}} \label{sec:instantft}
{\InstantFT} achieves high fine-tuning accuracy as \{FT, LoRA\}-All, while maintaining low backward FLOPs and memory usage similar to \{FT, LoRA\}-Last, thereby greatly improving the resource-accuracy trade-off.
The algorithm flow is illustrated in Fig. \ref{fig:instantft-flow}.

\subsection{Forward Pass} \label{sec:instantft-forward}
Unlike the baselines, {\InstantFT} introduces LoRA modules that connect each intermediate layer directly to the final layer, where we denote learnable parameters as $\left\{ \mathbf{A}^{i, L}, \mathbf{B}^{i, L} \right\}_{i = 0, \ldots, L - 1}$ ($\mathbf{W}^{i, L} = \mathbf{A}^{i, L} \mathbf{B}^{i, L}$).
{\InstantFT} produces an output as follows:
\begin{align}
  \vb{x}^L &= \left( \mathbf{W}^L \vb{x}^{L - 1} + \vb{b}^L \right) +
    \sum_{i = 0}^{L - 1} \mathbf{B}^{i, L} \mathbf{A}^{i, L} \vb{x}^i
    = \hat{\vb{x}}^L + \sum_{i = 0}^{L - 1} \mathbf{\Delta x}^i. \label{eq:skip-lora}
\end{align}
The first term $\hat{\vb{x}}^L$ is an output of the pre-trained base network, while the rest $\mathbf{\Delta x}^i = \mathbf{B}^{i, L} \mathbf{A}^{i, L} \vb{x}^i$ are obtained using LoRA adapters.

Since the pre-trained network is fixed during fine-tuning, $\left\{ \vb{x}^1, \ldots, \vb{x}^{L - 1} \right\}$ and $\hat{\vb{x}}^L$ remain unchanged for the same input $\vb{x}^0$.
{\InstantFT} thus introduces a \textbf{Forward Cache} to store and reuse these intermediate results.
As depicted in Fig. \ref{fig:instantft-flow}, {\InstantFT} first checks that the entry for the input $\vb{x}^0$ (with an index $j$) is present in the cache.
If so, {\InstantFT} reuses the precomputed results and only performs the adapter part to compute $\sum_i \mathbf{\Delta x}^i$; otherwise, {\InstantFT} also runs a forward pass of the pre-trained network to obtain $\left\{ \vb{x}^i \right\}, \hat{\vb{x}}^L$ and stores them in the cache for reuse.
As such, {\InstantFT} performs a forward pass of the pre-trained network only once per input, and repeatedly uses LoRA adapters in later epochs.
For a fixed dataset, the intermediate results for all samples are computed and cached in the first epoch.
This significantly reduces the overall computational cost and training time, as the pre-trained network involves more computation than LoRA modules (e.g., 33.7--40.7x higher forward FLOPs in our case, Table \ref{tbl:comparison}).
Additionally, {\InstantFT} has 3.8--5.9\% less forward FLOPs than LoRA-All, due to a lowered output dimension of LoRA modules.

\subsection{Cache Quantization} \label{sec:instantft-cache-quant}
Since the Forward Cache stores forward activations for every input, its size grows with the dataset size, model depth, and output dimensions, resulting in a prohibitive memory overhead.
To mitigate this, {\InstantFT} adopts the 4-bit NormalFloat (NF4) quantization~\cite{TimDettmers23}, which is well-suited for normally distributed parameters and activations.
This reduces the cache size by approximately 8x compared to FP32.
Note that the cached data is dequantized before used in the forward pass.

\subsection{Backward Pass} \label{sec:instantft-backward}
In backpropagation, given an activation gradient $\vb{dx}^i$, the $i$-th layer computes gradients w.r.t. the parameters and input:
\begin{align}
  \left\{ \begin{array}{l}
    \vb{dW}^i = \vb{dx}^i (\vb{x}^{i - 1})^\top \\
    \vb{db}^i = \vb{dx}^i \\
    \vb{dx}^{i - 1} = (\mathbf{W}^i)^\top \vb{dx}^i.
  \end{array} \right. \label{eq:backprop}
\end{align}
In LoRA (with the main network frozen), gradients are propagated as:
\begin{align}
  \left\{ \begin{array}{l}
  \vb{dB}^{j, k} = \vb{dx}^k (\vb{h}^{j, k})^\top \\
  \vb{dh}^{j, k} = (\mathbf{B}^{j, k})^\top \vb{dx}^k \\
  \vb{dA}^{j, k} = \vb{dh}^{j, k} (\vb{x}^{j})^\top \\
  \vb{dx}^{j} \gets \vb{dx}^{j} + (\mathbf{A}^{j, k})^\top \vb{dh}^{j, k},
  \end{array} \right. \label{eq:backprop-lora}
\end{align}
where $\vb{h}^{j, k} = \mathbf{A}^{j, k} \vb{x}^{j}$.
Given an output gradient $\vb{dx}^L$, {\InstantFT} computes LoRA gradients $\left\{ \vb{dA}^{i, L}, \vb{dB}^{i, L} \right\}_{i = 0, \ldots, L - 1}$ following Eq. \ref{eq:backprop-lora}.
Unlike $\vb{dx}, \vb{dW}$, which require quadratic computational complexity $O(d^2)$, LoRA gradients $\vb{dA}, \vb{dB}$ have lower complexity of $O(dr)$ ($r \ll d$).
{\InstantFT} uses the same number of adapters as LoRA-All but involves 14.1--40.6x less backward FLOPs than \{FT, LoRA\}-All, since it neither computes $\vb{dx}, \vb{dW}$ nor propagates $\vb{dx}$ through the entire network.
Notably, {\InstantFT} fine-tunes 12.3--52.1x more parameters with only 1.28--1.52x higher memory footprint compared to \{FT, LoRA\}-Last.
{\InstantFT} is likely to achieve better fine-tuning performance since trainable parameters are distributed across the network.

\begin{figure*}
  \centering
  \includegraphics[keepaspectratio, width=0.8\linewidth]{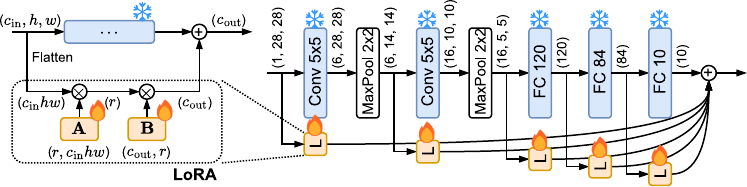}
  \caption{{\InstantFT} for LeNet-5-like model.}
  \label{fig:cnn-model}
\end{figure*}

\section{Implementations} \label{sec:impl}
We use a LeNet-5-like network~\cite{YannLecun98} (Fig. \ref{fig:cnn-model}) as a backbone, where five LoRA adapters ($r = 4$) are inserted into the Conv and FC layers for fine-tuning.
Each adapter takes an activation $\vb{x}^i \in \mathbb{R}^{c_\mathrm{in}}$ from the previous layer and computes an update $\vb{\Delta x}^i = \mathbf{B}^{i, L} \mathbf{A}^{i, L} \vb{x}^i \in \mathbb{R}^{c_\mathrm{out}}$ (Eq. \ref{eq:skip-lora}) to the last layer output $\hat{\vb{x}}^L$.
For Conv adapters, the input $\vb{x}^i \in \mathbb{R}^{c_\mathrm{in} \times h \times w}$ is first flattened into a vector $\vectorize(\vb{x}^i) \in \mathbb{R}^{c_\mathrm{in} hw}$ before computation (Fig. \ref{fig:cnn-model} (left)).

\begin{figure*}
  \centering
  \includegraphics[keepaspectratio, width=\linewidth]{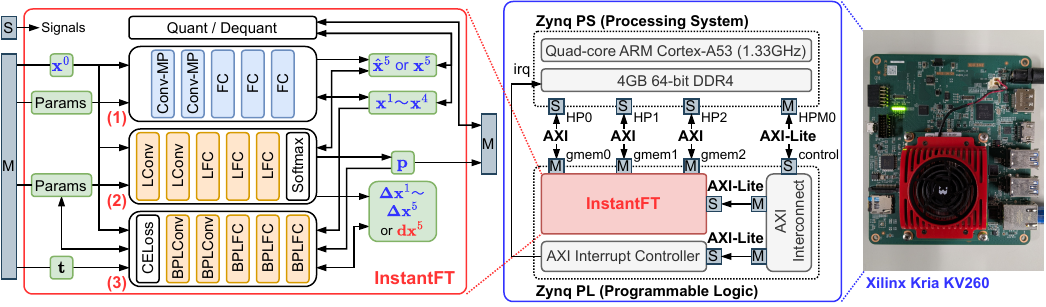}
  \caption{{\InstantFT} implemented on Xilinx Kria KV260.}
  \label{fig:board-level-design}
\end{figure*}

\begin{figure*}
  \centering
  \includegraphics[keepaspectratio, width=\linewidth]{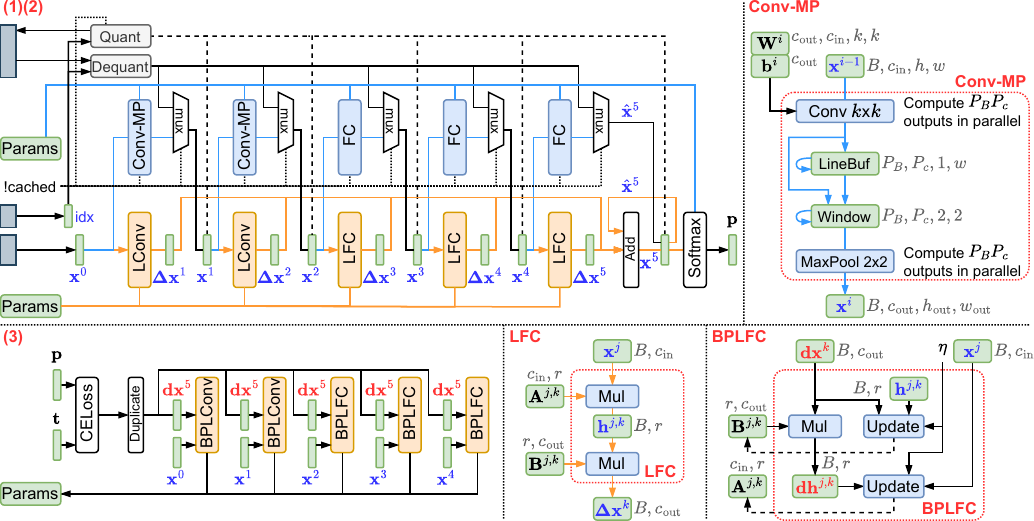}
  \caption{Block diagram of {\InstantFT} core.}
  \label{fig:block-detail}
\end{figure*}

\subsection{{\InstantFT} on FPGA} \label{sec:impl-hw}
We design a dedicated {\InstantFT} core for embedded FPGAs.
The core (Fig. \ref{fig:board-level-design}) consists of three module groups and a set of on-chip buffers (for storing inputs, parameters, forward activations, and gradients).
The first two groups (Fig. \ref{fig:block-detail}, top left) perform the forward pass of the base network and adapters, where \textbf{Conv-MP}/\textbf{FC} correspond to Conv-MaxPool/FC layers, and \textbf{LConv}/\textbf{LFC} are their respective LoRA adapters.
\textbf{Conv-MP} (Fig. \ref{fig:block-detail}, top right) consists of a pipeline of two modules: \textbf{Conv} performs 2D convolution and stores the output pixels into a line buffer (implemented as a shift register), while \textbf{MaxPool} fetches 2x2 pixels from this buffer and performs 2D max-pooling.

During fine-tuning, a mini-batch of samples and their corresponding indices are transferred to the core.
{\InstantFT} uses a Forward Cache placed on the external DRAM to avoid repeated forward passes of the base network for the same input.
The cached forward activations $\vb{x}^1, \ldots, \vb{x}^4, \hat{\vb{x}}^5$ are transferred to the on-chip buffers if available; otherwise, \textbf{Conv-MP}/\textbf{FC} compute them and send to the cache.
Then, \textbf{LConv}/\textbf{LFC} compute deltas $\vb{\Delta x}^1, \ldots, \vb{\Delta x}^5$ via matrix multiplications (Fig. \ref{fig:block-detail}, bottom center), which are summed together and added into $\hat{\vb{x}}^5$ to produce a final output $\vb{x}^5$.

\textbf{Dequant}/\textbf{Quant} handle the cache read/write operations.
\textbf{Dequant} reads 4-bit quantized cache entries for the input samples and writes them to respective buffers after dequantization, where indices are used to calculate memory addresses and locate the entries.
\textbf{Quant} performs the opposite.
The probabilities $\vb{p}$ for each label are obtained from the logits $\vb{x}^5$ via softmax; we adopt precomputed lookup tables~\cite{JavierDuarte18} instead of computing exponentials/inverses to simplify the logic.

Based on the predictions $\vb{p}$ and true labels $\vb{t}$, the last group (Fig. \ref{fig:block-detail}, bottom left) performs backpropagation and gradient descent to update adapters.
\textbf{CELoss} (cross-entropy loss) computes an output gradient $\vb{dx}^5$ and copies it to buffers that originally stored deltas $\vb{\Delta x}^i$.
Then, given $\vb{dx}^5$, saved activations $\vb{x}^j, \vb{h}^{j, k} = \vb{A}^{j, k} \vb{x}^j$, and a learning rate $\eta \in \mathbb{R}^+$, \textbf{BPLConv}/\textbf{BPLFC} (Fig. \ref{fig:block-detail}, bottom right) perform matrix operations to obtain parameter gradients $\vb{dA}, \vb{dB}$ and update LoRA parameters via SGD ($\vb{A} \gets \vb{A} - \eta \vb{dA}$).

To improve latency, these modules are parallelized through loop unrolling and buffer partitioning (e.g., \textbf{Conv-MP} computes output pixels for multiple samples and channels per clock cycle).
Besides, the LoRA modules (\textbf{LConv}/\textbf{LFC} and \textbf{BPLConv}/\textbf{BPLFC}) are executed in parallel to simultaneously compute deltas/gradients and update parameters at once.
Importantly, {\InstantFT} allows such an optimization because all LoRA adapters directly connect to the last layer.
In contrast, LoRA-All requires LoRA adapters to be executed sequentially, as the gradient $\vb{dx}$ should be backpropagated through the entire network.

We use Xilinx Kria KV260 running PetaLinux 2022.1 as a target platform (Fig. \ref{fig:board-level-design}, right), which features a Zynq UltraScale+ MPSoC FPGA, a quad-core ARM Cortex-A53 CPU (1333MHz), and 4GB DDR4 DRAM.
We use Xilinx Vitis 2024.1 to implement the core in C/C++ and generate an FPGA bitstream.
The core along with two AXI IPs (an interrupt controller and interconnect) are clocked at 200MHz and implemented on PL (Programmable Logic).
The core has 128-bit AXI manager interfaces connected to the HP (high-performance) ports for data transfer from/to the DRAM on PS (Processing System), and a 32-bit AXI-Lite subordinate interface for control register accesses.

The frozen network parameters and initial LoRA parameters are pre-loaded into on-chip buffers to minimize the DRAM accesses.
We use Q8.16 fixed-point format to represent forward activations, and Q4.12 for parameters and gradients.
The host code is developed in C/C++ using XRT (Xilinx Runtime) library to communicate with the core.
It transfers input data (samples/labels/indices) and the updated LoRA parameters via memory-mapped AXI interfaces.

\begin{figure*}
  \begin{minipage}[b]{\hsize}
    \centering
    \includegraphics[keepaspectratio, width=\linewidth]{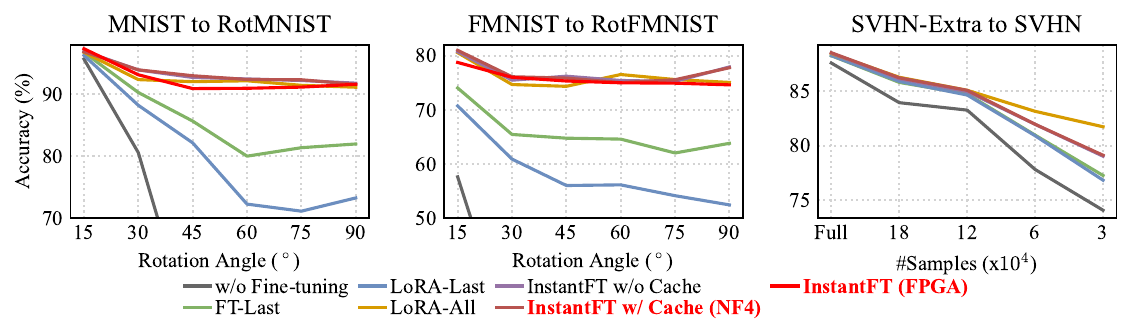}
    \subcaption{Accuracy.}
    \label{fig:eval-results-acc}
  \end{minipage}
  \begin{minipage}[b]{0.66\hsize}
    \centering
    \includegraphics[keepaspectratio, width=\linewidth]{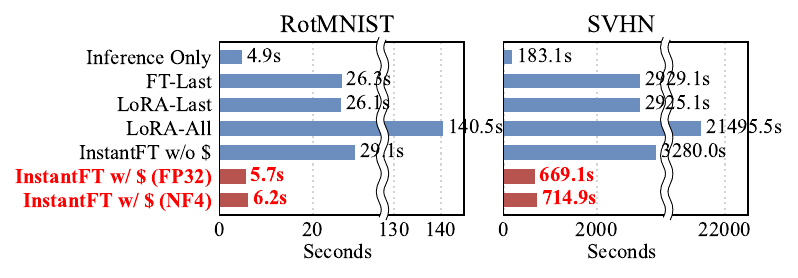}
    \subcaption{Execution time.}
    \label{fig:eval-results-time}
  \end{minipage}
  \begin{minipage}[b]{0.33\hsize}
    \centering
    \includegraphics[keepaspectratio, width=\linewidth]{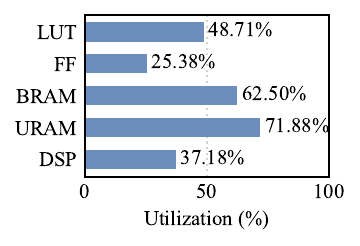}
    \subcaption{FPGA resource utilization.}
    \label{fig:eval-results-res}
  \end{minipage}
  \caption{Evaluation of {\InstantFT} and baselines.}
  \label{fig:eval-results}
\end{figure*}

\section{Experimental Results} \label{sec:eval}
\subsection{Experimental Setup} \label{sec:eval-setup}
We compare {\InstantFT} with the baseline fine-tuning approaches (Fig. \ref{fig:finetune-approaches}) under the following three cases:
\begin{enumerate}[leftmargin=*, label=(\arabic*)]
  \item The LeNet-5-like model (Fig. \ref{fig:cnn-model}) is pre-trained on MNIST and then fine-tuned on its rotated version (\textbf{RotMNIST}). \label{enum:eval-case1}
  \item Pre-trained on Fashion-MNIST (FMNIST) and fine-tuned on its rotated counterpart (\textbf{RotFMNIST}). \label{enum:eval-case2}
  \item Pre-trained on SVHN-Extra and fine-tuned on the test split of SVHN. \label{enum:eval-case3}
\end{enumerate}
We randomly pick 1024 samples each from the test sets of \{MNIST, FMNIST\} and rotate them by $\theta$: 15$^\circ$, 30$^\circ$, $\ldots$, 90$^\circ$ to generate six variants of Rot\{MNIST, FMNIST\} datasets.
In \ref{enum:eval-case3}, we pre-train the network with different numbers of samples $\left| \mathcal{D} \right|$: 30000, 60000, 120000, 180000, 521131.
We simulate challenging fine-tuning tasks with greater data distribution shifts using a larger $\theta$ or smaller $\left| \mathcal{D} \right|$.
The model is pre-trained and fine-tuned for 10 epochs with a batch size of 20.
The learning rate $\eta$ is set to 0.1 and 0.025 in \ref{enum:eval-case1}--\ref{enum:eval-case2} and \ref{enum:eval-case3}.
The code for pre-training and fine-tuning is developed in C and compiled using GCC 11.2.0 with \texttt{-O3} flag.
We use OpenMP to utilize all four CPU cores.

\subsection{Accuracy} \label{sec:eval-accuracy}
Fig. \ref{fig:eval-results-acc} shows the classification accuracy of {\InstantFT} and baselines on three cases.
We report the average of ten runs with different random seeds.
Without fine-tuning, the accuracy drops sharply under challenging settings (larger rotation angles (Rot\{MNIST, FMNIST\}) or fewer pre-training samples (SVHN)).
While fine-tuning the last layer recovers the accuracy to some extent, FT-Last/LoRA-Last is still 0.04--12.2\%/0.5--20.3\% less accurate on RotMNIST (6.5--13.6\%/9.8--22.6\%, 0.2--4.5\%/0.2--4.9\% less accurate on RotFMNIST, SVHN) compared to fine-tuning the entire network with LoRA adapters (LoRA-All).
In contrast, {\InstantFT} performs comparably or even better than LoRA-All; it maintains an accuracy of 91.8/77.9/79.0\% on three cases under the most difficult settings ($\theta = 90^\circ, \left| \mathcal{D} \right| = 30000$), while LoRA-All is 91.2/75.1/81.7\% accurate.
{\InstantFT} shows a similar robustness to data distribution shifts as LoRA-All, with an accuracy drop of 5.2/2.8/9.4\% (LoRA-All: 5.7/6.2/6.7\%).
Besides, NF4 quantization of the forward cache only leads to a marginal accuracy loss of 0.09--0.4\%.
The FPGA implementation (\textcolor{red}{red}) attains accuracy within 3.2\% of its software counterpart and remains robust under data distribution shifts.

\subsection{Execution Time} \label{sec:eval-time}
Fig. \ref{fig:eval-results-time} shows the time required for 10-epoch fine-tuning on RotMNIST and SVHN, measured on an ARM Cortex-A53 CPU.
Due to higher forward/backward FLOPs, LoRA-All takes 5.3x more time to fine-tune all Conv/FC layers on RotMNIST compared to only updating the last layer (FT-Last/LoRA-Last).
While {\InstantFT} employs the same number of LoRA adapters, it runs 4.8x faster than LoRA-All on RotMNIST, as it obviates the need to compute output gradients $\vb{dx}^i$ (except $\vb{dx}^5$) and saves the computational cost for backpropagation by 20.9x.
Forward Cache leads to further 5.1x speedup on RotMNIST, as forward activations of the base network are computed only once per input at the first epoch and then reused in the subsequent epochs.
Notably, {\InstantFT} performs fine-tuning with only a 17\% increase in execution time compared to the inference-only case, and successfully addresses the discrepancy between training and testing data.
While NF4 (de)quantization of cached activations introduces an additional 8.8\% overhead, {\InstantFT} still runs significantly faster than baselines.
The results on SVHN show a similar trend except for the 38--153x longer execution time, which is mainly due to 72x larger dataset size (73257/1024 images).

The FPGA implementation of {\InstantFT} (with quantized cache) only takes 0.36s on Kria KV260 to fine-tune the model for 10 epochs, which is 17.4x faster than the CPU counterpart, thereby enabling on-the-fly adaptation of CNNs.
For a fair comparison, the execution time includes various overheads such as PS--PL data transfers and FPGA kernel invocations.

\subsection{Power Consumption} \label{sec:eval-power}
During fine-tuning, we measure the power consumption of KV260 every 50ms using an onboard INA260 sensor and compute its average value.
While the FPGA-implemented {\InstantFT} consumes 0.24W more power (3.79W/4.03W) compared to only using ARM Cortex-A53, it achieves 16.32x higher energy-efficiency thanks to the attained speedup.

\subsection{Forward Cache Size} \label{sec:eval-cache-size}
The size of Forward Cache depends on both the number of samples in a given dataset and model architecture.
On Rot\{MNIST, FMNIST\} and SVHN, the full-precision cache requires 7.33/524.52MB of memory, which is brought down to 1.02/72.89MB by NF4 quantization (7.20x reduction) without compromising accuracy (Fig. \ref{fig:eval-results-acc}).

\subsection{FPGA Resource Utilization} \label{sec:eval-fpga-resource}
Fig. \ref{fig:eval-results-res} shows the FPGA resource utilization of {\InstantFT} on Kria KV260.
The design effectively uses on-chip memory resources to store (frozen/LoRA) parameters, activations, and gradients, thereby minimizing external memory accesses and improving overall performance.
The on-chip memory can be further saved by applying low-bit quantization to parameters and gradients as well.
In addition to LeNet-5, larger-scale networks can be fine-tuned on KV260, thanks to the low utilization of logic resources (LUTs/DSPs).

\section{Summary} \label{sec:summary}
This paper presents {\InstantFT}, a new ultra-fast fine-tuning method that enables edge applications to quickly adapt to data distribution shifts within subseconds.
{\InstantFT} leverages trainable adapters directly connected to the output layer, significantly saving the computational cost of backpropagation.
In addition, it introduces 4-bit quantized Forward Cache to eliminate redundant forward passes of the frozen base network.
The proposed FPGA-based design fully exploits the high parallelism and simplified computation flow of the adapters.
Experimental results demonstrate that {\InstantFT} achieves accuracy comparable to the original LoRA in just 0.36s on Xilinx Kria KV260, while being 17.4x faster and 16.3x more energy-efficient than ARM Cortex-A53.
While {\InstantFT} is currently evaluated on small-scale networks, we aim to extend its application to large-scale models including LLMs.

\renewcommand{\baselinestretch}{1.0}
\bibliographystyle{unsrt}

\input{refer.tex}

\end{document}

%% file: settings.tex
\usepackage{arxiv}

\usepackage{graphicx}

\usepackage[utf8]{inputenc}
\usepackage[T1]{fontenc}

\usepackage{amsmath}
\usepackage{amssymb}
\usepackage{amsfonts}
\usepackage{amsthm}
\usepackage{color}

\usepackage{cite}
\usepackage{hyperref}
\usepackage{url}

\usepackage{enumitem}

\usepackage{physics}

\usepackage{makecell}
\usepackage{multicol}
\usepackage{multirow}

\usepackage{subcaption}
\usepackage{siunitx}

\expandafter\def\expandafter\UrlBreaks\expandafter{\UrlBreaks\do\/
  \do\*\do\-\do\~\do\'\do\"\do\-}

\DeclareMathOperator{\vectorize}{\mathrm{vec}}

\def\BibTeX{{\rm B\kern-.05em{\sc i\kern-.025em b}\kern-.08em
  T\kern-.1667em\lower.7ex\hbox{E}\kern-.125emX}}

\newcommand{\InstantFT}{InstantFT}